\documentclass[runningheads]{llncs}

\usepackage{eccv}

\usepackage{algorithm} 
\usepackage{algpseudocode} 

\usepackage{eccvabbrv}

\usepackage{graphicx}
\usepackage{booktabs}

\usepackage[accsupp]{axessibility}  

\usepackage[pagebackref,breaklinks,colorlinks,citecolor=eccvblue]{hyperref}

\usepackage{orcidlink}

\begin{document}

\title{VisionGPT-3D: A Generalized Multimodal Agent for Enhanced 3D Vision Understanding} 

\titlerunning{VisionGPT-3D}

\author{%
\begin{tabular}{ccc}
Chris Kelly\thanks{Authors contributed equally.} & Luhui Hu\textsuperscript{$\star$} & Jiayin Hu\textsuperscript{$\star$} \\
\textit{Stanford University} & \textit{Seeking AI} & \textit{University of California, Los Angeles} \\
\email{ckelly24@stanford.edu} & & \email{greenlake@ucla.edu} \\[2ex]
Yu Tian & Deshun Yang & Bang Yang \\
\textit{Harvard University} & \textit{Seeking AI} & \textit{Peking University} \\
\email{ytian11@meei.harvard.edu} & & \email{yangbang@pku.edu.cn} \\[2ex]
Cindy Yang & Zihao Li & Zaoshan Huang \\
\textit{University of Washington, Seattle} & \textit{Seeking AI} & \textit{Seeking AI} \\ \email{c1ndyy@uw.edu}
 & \email{li981354@seeking.ai} & \\[2ex]
 & Yuexian Zou & \\
 & \textit{Peking University} & \\
 & \email{zouyx@pku.edu.cn} &
\end{tabular}
}

\institute{} 

\authorrunning{C. Kelly, L. Hu, J. Hu et al.}

\maketitle

\begin{abstract}
The evolution of text to visual components facilitates people's daily lives, such as generating image, videos from text and identifying the desired elements within the images. Computer vision models involving the multimodal abilities in the previous days are focused on image detection, classification based on well-defined objects. Large language models (LLMs) introduces the transformation from nature language to visual objects, which present the visual layout for text contexts. OpenAI GPT-4 has emerged as the pinnacle in LLMs, while the computer vision (CV) domain boasts a plethora of state-of-the-art (SOTA) models and algorithms to convert 2D images to their 3D representations. However, the mismatching between the algorithms with the problem could lead to undesired results. In response to this challenge, we propose an unified VisionGPT-3D framework to consolidate the state-of-the-art vision models, thereby facilitating the development of vision-oriented AI. VisionGPT-3D provides a versatile multimodal framework building upon the strengths of multimodal foundation models. It seamlessly integrates various SOTA vision models and brings the automation in the selection of SOTA vision models, identifies the suitable 3D mesh creation algorithms corresponding to 2D depth maps analysis, generates optimal results based on diverse multimodal inputs such as text prompts.
\end{abstract}

\section{Introduction}
\label{sec:intro}
Human-being feel it’s easy to describe a vivid story by language and make other people outline a lively fantasy in the mind. It could be hard for AI to generate visual context from its corresponding text. Although models like GPT-4 have established a remarkable benchmark for LLMs, and SORA demonstrate the extension of generating 3D visual components, the field of multimodal CV models is an ever-changing frontier, full of potential. \\
\\
Our VisionGPT-3D integrates multiple large-scale models such as SAM (segment anything model) which has the ability to segment or cut out object within an image, the YOLO (You only look once) models which can detect objects in the image, DINO (distillation of image representation through self-attention) which is a self-supervised learning framework proposed for image representation learning. Each model has its pros and cons. DINO can capture the relationships between different parts of the image and get good performance on fine-grained object detection. SAM focuses on spatial attention to emphasize important regions of image. YOLO demonstrate strong performance in object detection benchmarks. Self-supervised models like DINO can be computationally expensive. SAM may not address model performance. YOLO may not perform as well on fine-grained object detection tasks.  The choice between DINO, SAM and YOLO may depend on the specific tasks such as self-supervised learning, spatial attention or real-time object detection. \\
\\
VisionGPT-3D creates the optimized solution with the joint of multiple models or makes selection of the models based on the task types. The paper focuses on integrating reconstructing 3D image from its 2D representation into VisionGPT-3D by exploring multiple techniques, such as multi-view stereo creates a dense point cloud from the depth map followed by a surface mesh created from the dense point cloud to form triangles, which results in a representation of the scene’s geometry; structure from motion estimates the 3D structure of a scene from a set of 2D images taken from different viewpoints; Depth from stereo compares disparities between corresponding points in stereo image pairs, which is effective for close-range objects; Photometric stereo estimates depth from variations in image intensity due to changes in illumination direction, which is effective for surfaces with varying reflectance but requires controlled lighting; Light detection and ranging users laser light to measure distances providing high precision for outdoor scenes but requires expensive hardware; Time of flight cameras measures the time it takes for light to travel to the object and back gives real-time depth sensing but it is less accurate for shiny surfaces. To implement the feature into VisionGPT-3D framework, we explore the steps (1) obtain the depth map from the image. (2) create point cloud. (3) generate mesh from point cloud. (4) create video from context. 

\section{Method}

\subsection{Generate depth map}
Depth map provides information about the distance or depth of objects in a scene relative to a viewer or a camera. Figure 1 shows an example of depth map representing a hare. 
\begin{figure}
    \centering
    \includegraphics[width=0.5\linewidth]{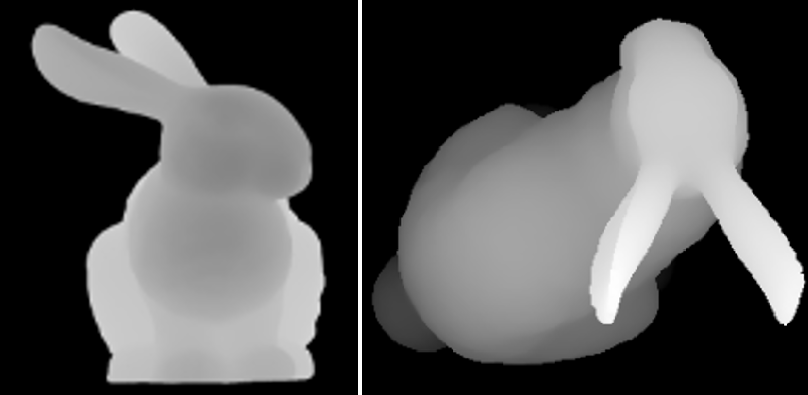}
    \caption{Image depth map}
    \label{fig:figure1}
\end{figure} 

It is a representation of the 3D structure of a scene captured in a 2D image. Each pixel in the depth map corresponds to a specific point in the scene. The pixel intensity represents the depth of that point. High pixel values correspond to objects that are farther away while lower values correspond to closer objects. Depth map can be generated from the disparities between corresponding points in the images taken from different positions. It can also be estimated by scene understanding neural network such as monocular depth estimation in real-time with adaptive sampling, DepthNet, MonoDepth models. Transfer learning and pretrained models are also used for depth estimation. By leveraging deep learning models pretrained for related tasks and fine-tuning them for depth estimation. Figure 2 below show the depth information of the images taken from different angles.

\begin{figure}
    \centering
    \includegraphics[width=0.5\linewidth]{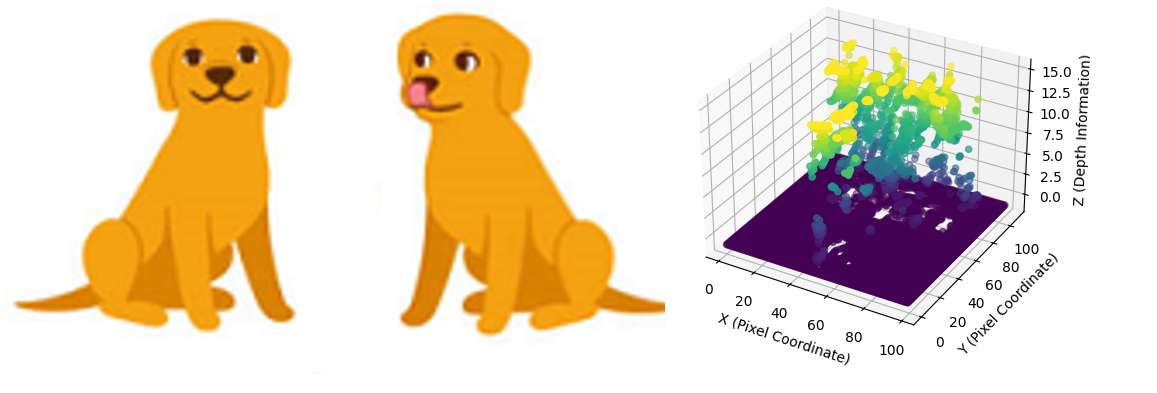}
    \caption{Depth of 2D images from different angles}
    \label{fig:figure2}
\end{figure}
It requires multiple images to get the depth map. Sometimes, we don’t have many images representing an object. In order to estimate depth map from a single 2D image, MiDas(Monocular Depth Estimation in Real-Time with Adaptive Sampling) is trained on a large dataset with corresponding depth information. It learns the associate visual features in an image with their depth values. MiDaS typically leverages convolutional neural network for depth estimation. It is trained on a dataset containing pairs of monocular images and their depth maps. The model learns to predict the depth by minimizing the difference between predictions and the ground truth depth values in the training. MiDas uses adaptive sampling to allocate resource based on image structures. The intricate structures get more resources than the simpler regions. The complex structures could be the farthest part in the image yield to be the less important part. The paper proposes a new sampling method which focuses on the key points of an image. The light green circles in figure 3 shows the key points in the images. 
\begin{figure}
    \centering
    \includegraphics[width=0.5\linewidth]{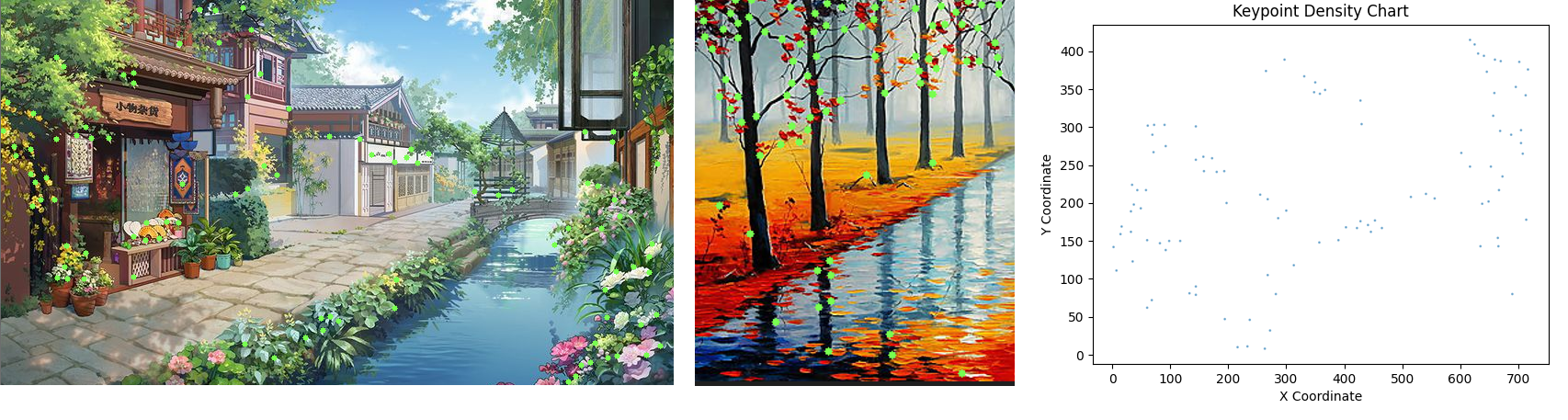}
    \caption{2D image key points}
    \label{fig:figure3}
\end{figure}

The sampling algorithms allocate the resource based on the density of the key points. The algorithm describes below. 

\begin{algorithm}
	\caption{Resource allocation based on key points} 
	\begin{algorithmic}[1]
        \State Set up K-Means cluster
		\For {$2D-Images$}
			\State Group the key point into K clusters based on its coordinator
		\EndFor
  	\State Calculate the number of the key points in each group
        \State Allocate resource based on the number of key points in each group
	\end{algorithmic} 
\end{algorithm}

Pretrained MiDaS model allows users to generate depth map without training from scratch. To get better model prediction precision performance when process various image and scenes, we modify and fine-tune the model based on the requirement and minimize the efforts brought to the users.

\subsection{Create point cloud from depth map}

Point clouds serve as a fundamental step in 3D reconstruction pipelines. It presents a detailed description of the structure and layout of a scene. By converting 2D pixel coordinates to 3D ray in camera space with intrinsics and associate the depth value from depth map with the corresponding pixel, the process converts the 2D-3D association to world coordinates using extrinsic finally. The point cloud is saved in a common format file as like PLY and is visualized by 3D visualization tools such as Matplotlib, Open3D. The point cloud can be generated from depth map by using Open3D library. In order to generalize the point clouds from 2D depth maps, a neural network takes the depth map as input and outputs predicted 3D coordinates for each pixel. The neural network consists of components such as analyze key depth regions, identify object boundaries, filter noise, object segment, compute surface normal, analyze depth gradients. Filter out the noise before process the image can improve the learning result. The noise may affect the subsequent processing steps. In order to understand the scene and identify key depth regions, it’s necessary to check the range of the depth values and analyze the distribution of depth values. By finding out the key depth regions, it’s efficient to place corresponding resources based on the density of key depth regions. Identify the object boundary helps to group the key depth regions belonging to an object for coherence learning.  The sharp changes in depth often correspond to object boundaries or occlusions. Object segmentation search the meaningful objects or regions and partition the image based on its structure. It allows program to perceive the surroundings and make informed decisions. The VisionGPT-3D can respond the object based on depth. It allows for selective manipulation of specific objects within the scene. It is vital for collision avoidance when VisionGPT-3D is used for autonomous vehicles, drones or industry assemblies. Knowing the boundaries of objects in depth helps in path planning and obstacle avoidance. Surface normals provides a concise representation of the local geometry of a surface and indicate the direction perpendicular to the surface at a specific point. They are used for understanding the shape and orientation of surfaces within a 3D scene. They are essential for feature extraction and clustering in point cloud processing. They help distinguish between objects with different geometric properties, aiding in classification and identification tasks. Depth gradients provide information about the rate of change of depth values in an image. They help identify depth discontinuities and object boundaries, contributing to more accurate segmentation. They are useful for 3D object recognition and labeling. In the previous context, we proposed to convert the labeled 2D image into its 3D representation as the respond for users in VisionGPT-3D. We further labeled the object in 3D image as the training dataset to train the model to generate related 3D images when the users wish to have other answers or styles responses. The overall network architecture describes in Figure 4 below. 
\begin{figure}
    \centering
    \includegraphics[width=0.5\linewidth]{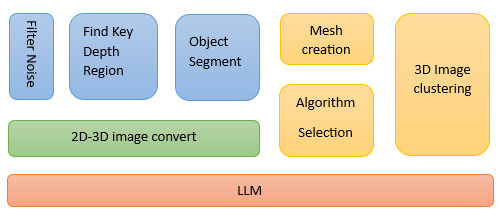}
    \caption{AI based 2D-3D image converter pipeline}
    \label{fig:figure4}
\end{figure}

In this paper, we focus on innovation of object segmentation within 3D image. There are many existing algorithms and methods can be used for object segmentation in depth map. Analyzing depth map accurately can help to generate a successful point cloud. Efficiently object segmentation leads to a better understanding of scene with the depth map. Object segmentation algorithms such as thresholding, watershed segmentation, graph-cut segmentation, mean-shift segmentation, superpixel-based segmentation, region growing are common in computer vision processing. The choice of segmentation algorithms depends on the specific characteristics of the depth map, the complexity of the scene and the desired level of segmentation details. Thresholding is a basic technique where the regions are separated by a certain threshold. Region growing starts with a seed pixel and adds neighboring pixels to the region if their depth values are similar. The seed selection could impact the algorithms performance and result. Watershed segmentation separates different objects based on depth discontinuities by assigning pixels to catchment basins. It can lead to over-segmentation in smooth areas or under-segmentation in the presence of weak intensity differences. Graph-cut segmentation construct a graph representing the depth map and apply graph-cut algorithms to find optimal segmentation. The computation complexity can be expensive especially for large dataset or high-dimensional feature spaces. Graph-cut methods often rely on pixel-wise information, and regions with similar appearance may not be effectively separated. Mean-shift clustering groups pixels with similar depth values and forms distinct objects. It can potentially lead to the loss of details after merge regions with similar color while the areas have different regions having homogeneous color. Superpixel-based segmentation decompose the depth map into compact image regions. It uses clustering method to generate superpixel based on depth information. It reduces computational complexity but can lead to excessive segmentation in complex areas. It’s inefficient to manually select the algorithms and segment the image. We propose an AI based approach to select the algorithms for object segmentation based on the image characteristics.  The input is the depth image and output are the optimized segmentation algorithms with the result. The training minimizes the errors between ground truth algorithms selection with the model output based on the same image. It employs convolutional network as the base architecture to process the image and include the ranking method to return top k corresponding algorithms, apply the top relative algorithms to segment the image. The model is able to scale up and process large datasets. With this approach, the object segmentation efficient and correctness are improved. The point cloud can be generated based on the segmented depth map by tools such as Open3D or OpenCV. 

\subsection{Generate mesh from point cloud}
Generating mesh from a point cloud involves creating a surface representation by connecting the points with triangles. The common algorithms used to generate a mesh from a point cloud including Delaunay Triangulation, surface reconstruction, alpha shapes, marching cubes, ball pivoting algorithms, power crust, screened Poisson reconstruction, ball and triangle growing, depth map-based techniques, Poisson surface reconstruction, greedy projection triangulation, Voronoi diagram-based meshing. Delaunay triangulation maximizes the minimum angle of all triangles in the mesh to reduce numerical computations caused by small triangles and improves the computation efficiency. The defect is it can produce flat and long triangles in the areas with irregular point distributions.  Alpha shapes can create concave hulls when deal with point clouds representing objects with non-convex shapes. However, it depends on the selection of parameter alpha. The alpha shape computation can become a limitation for complex shapes. Poisson surface reconstruction is based on a Poisson equation to reconstruct a scalar field representing the underlying surface. It tends to generate smooth and continuous surfaces and is hard to capture sharp features in geometry. The gradient and Laplacian operators of Poisson equation in 3D are defined as like below. The Poisson equation is modified to an optimized problem by defining an object function that minimizes the difference between the Laplacian of the scalar field and the density function.The density function is constructed based on the input point cloud. Based on the formulas as like below, it tries to calculate the partial derivatives of the scalar field which requires the surface to be smooth. The VisionGPT-3D selects the corresponding algorithms based on the analysis of algorithms definition improving accuracy.
\begin{figure}
    \centering
    \includegraphics[width=0.5\linewidth]{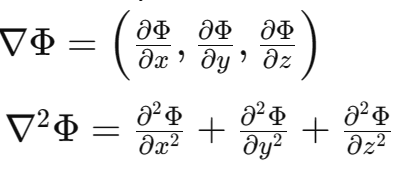}
\end{figure}
While moving least square is a local approximation method fitting a polynomial surface which is more capable of preserving sharp features in the reconstructed surfaces. MLS can be more efficient than Poisson surface reconstruction as it employs the approximation method. In the previous section, we can find the object edges from sharp change in pixel values by analyze depth gradient within depth map. It helps us to select suitable algorithms to deal with smooth or sharp surface in mesh creation. Ball pivoting algorithm and ball growing algorithm connect neighboring balls to form triangles. Ball pivoting algorithm can struggle to adapt to surfaces with varying curvatures while ball growing algorithm is well-suited for surfaces with varying curvature. However, Ball pivoting algorithm has better performance than the other one. They have different connecting points and forming triangles causing the different adaptability to handle varying curvatures. The fixed size ball in ball pivoting can be a limitation to deal with varying curvatures. Ball growing algorithm can adapt to regions with different point densities, ensuring a better surface reconstruction. In the previous section, we are able to identify the point densities by identifying the key points or analyze the pixel values within the depth map. The depth map analysis result is used for adapting an optimized mesh creation algorithm for a certain region within an image. 

\subsection{Validate mesh correctness}
Validating the correctness of a mesh generation algorithm is a crucial step to ensure that the resulting mesh accurately represents the underlying geometry. Generally, people inspect the generated mesh using 3D visualization tools. Check for smoothness, continuity and whether if captures the important features of the underlying geometry. The method provides an intuitive understanding but may not be sufficient for quantitative assessment. Such as the incorrect generating result in figure 5. 
\begin{figure}
    \centering
    \includegraphics[width=0.5\linewidth]{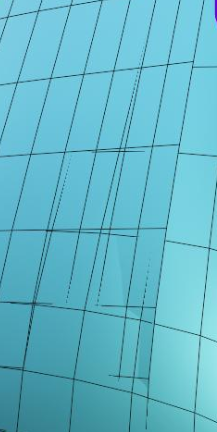}
    \caption{Incorrect generated mesh}
    \label{fig:figure5}
\end{figure}
The VisionGPT-3D validates the generated mesh by model trained from comparative analysis. It regenerates the mesh when the previous result fails. Based on the common algorithms used in mesh validation, VisionGPT-3D selects an optimized validate method to validate the result. Besides the visual inspection, there are other validation methods such as surface deviation analysis, edge length analysis, volume conservation, mesh quality metrics, convergence analysis. Surface deviation analysis compares the generated mesh with a reference model or ground truth by computing the surface deviation at corresponding points. Metrics like Hausdorff distance or point-to-surface distance can be used to quantify deviations. Smaller deviations indicate better accuracy. Edge Length Analysis analyzes the distribution of edge lengths in the generated mesh. It helps ensure that the mesh is refined appropriately in areas of high curvature or detail. Volume Conservation ensures that the volume enclosed by the mesh is consistent with the original object's volume. The validation methods have the similar factors that comparing the ground truth with the errant result to identify the correctness. 

\subsection{Generate video from static frames}
Sometimes, people may describe a context representing a consecutive scenario. As an example, a falcon is flying in the snowing sky. Generating a video from a 3D image involves visualizing the image data across different frames to create the illusion of motion. It's not hard the frames are all presented. We could use library like ffmpeg to generate them. The steps are (1) create a fiture and 3D axes. (2) set up the initial 3D image. (3) update the 3D plot for each frame. (4) set the number of frames and create the animation. It could be more complex if we don't have all the frames for the video. We use an alternative approach to do that. Instead of being focus on the frames, we places and routes the objects onto the scenes in the runtime. The detail process describes as follows. The language understanding model segments the words firstly. It tries to find out the words representing the scenes or objects in the scenes. In our previous analysis, we know the objects inside images can be analyzed by their depth and we understand the collisions among them. We are able to label the image to differentiate the scene from the object which will collide with the scene. We could also filter the image into multiple levels regarding the objects we observed. Based on these information, we're able to place the objects onto the scenes considering the collisions. When the VisionGPT-3D places and routes the objects, it tries to make the object moving based on basic physical laws. In the common approaches, the models are trained based on statistics and places the objects based on experiences. However, the statistics might not describe the object movement correctly. In our innovation, we generate the moving path in the runtime based on the collision information obtained from image. 

\subsection{Validate generate video correctness}
Validation is an vital step in the visual context processing as the bad visual components leaves a more directly negative influence in people's minds than other text based components. As showed in Figure 6, the temples in the 
farthest positions inside the image are generated in an incorrect size ratio comparing to the background scene and depth of the front object.
\begin{figure}
    \centering
    \includegraphics[width=0.5\linewidth]{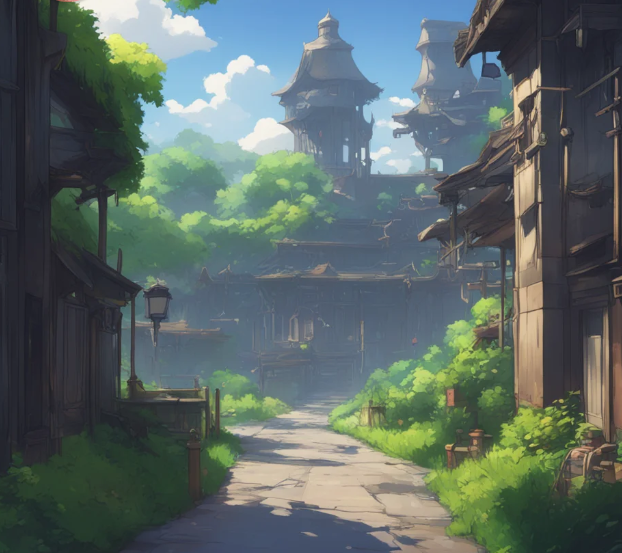}
    \caption{Incorrect object inside the image}
    \label{fig:figure6}
\end{figure}

Figure 7 shows an incorrect generate image. 
\begin{figure}
    \centering
    \includegraphics[width=0.5\linewidth]{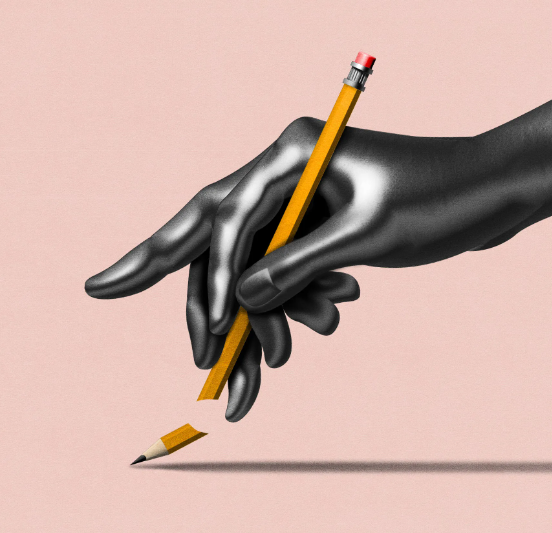}
    \caption{Incorrect image}
    \label{fig:figure7}
\end{figure}

While in the validation process, the framework will need to check color accuracy, validate that colors in the video are accurate and consistent; Frame Rate, confirm that the video plays at the intended frame rate and other validation points as the features for machine learning model training.

\section{Summary}

In this paper, we proposed an unified visionGPT-3D framework can select optimized mesh creation, depth map analysis algorithms based on machine learning model to predict the users' requirements. Combining the traditional vision processing methods with AI models into the unified system to maximize the capability of visual application transformations. By leveraging the self-supervised learning to train the model and select the well-suited algorithms in each step of 3D reconstruction from 2D image. During the work, we find the limitation while working in non-GPU environment while some libraries are not available or with low performance. In order to further reduce the model training cost and improve its efficiency, prediction precision, we plan to work on algorithms optimization based on a self-design low cost generalized chipset.

\clearpage  


%
%

\bibliographystyle{splncs04}

\end{document}